\documentclass{article}
\usepackage{spconf,amsmath,graphicx}

\usepackage{cite}
\usepackage{CJK}
\usepackage{amsmath,amssymb,amsfonts}
\usepackage{graphicx}
\usepackage{textcomp}
\usepackage{xcolor}
\usepackage{multirow}
\usepackage[linesnumbered,ruled,vlined]{algorithm2e}
\usepackage{subfigure}
\usepackage{bbding}
\usepackage{algpseudocode}
\usepackage{amsmath}
\usepackage{booktabs}
\usepackage{svg}
\usepackage{hyperref}
\usepackage{url}
\usepackage{pifont}
\usepackage{diagbox}

\hypersetup{hidelinks}
\usepackage[numbers,sort&compress]{natbib}
\def\BibTeX{{\rm B\kern-.05em{\sc i\kern-.025em b}\kern-.08em
    T\kern-.1667em\lower.7ex\hbox{E}\kern-.125emX}}

\newcommand{\tabincell}[2]{\begin{tabuzlar}{@{}#1@{}}#2\end{tabular}}

\title{A novel efficient Multi-view traffic-related object detection framework}
\name{
    Kun Yang$^1$,
    Jing Liu$^1$,
    Dingkang Yang$^1$,
    Hanqi Wang$^1$,
    Peng Sun$^2$,
    Yanni Zhang$^3$,
    Yan Liu$^4$,
    Liang Song$^{1,3*}$\thanks{* Corresponding author.}\thanks{This work is partially supported by the Shanghai Key Research Laboratory of NSAI and NSFC Grant 62250410368.}
}
\address{
    \textsuperscript{1}Academy for Engineering \& Technology, Fudan University, China \\
    \textsuperscript{2}Duke Kunshan University, China \\
    \textsuperscript{3}Shanghai East-bund Research Institute on NSAI, China \\
    \textsuperscript{4}Jiangxi Open University, China \\
}
\begin{document}
%\ninept
%
\maketitle
\begin{abstract}
With the rapid development of intelligent transportation system applications, a tremendous amount of multi-view video data has emerged to enhance vehicle perception. However, performing video analytics efficiently by exploiting the spatial-temporal redundancy from video data remains challenging. Accordingly, we propose a novel traffic-related framework named CEVAS to achieve efficient object detection using multi-view video data. Briefly, a fine-grained input filtering policy is introduced to produce a reasonable region of interest from the captured images. Also, we design a sharing object manager to manage the information of objects with spatial redundancy and share their results with other vehicles. We further derive a content-aware model selection policy to select detection methods adaptively. Experimental results show that our framework significantly reduces response latency while achieving the same detection accuracy as the state-of-the-art methods.
\end{abstract}
\vspace{-0.3cm}
\section{Introduction}
\label{sec:intro}
\vspace{-0.3cm}
Recently, video analytics-based applications have been widely applied to support smart cities, including traffic monitoring~\cite{guo2021crossroi,liu2022appearance,liu2022collaborative} and video surveillance~\cite{jain2020spatula,liu2022learning}. To improve the efficiency of video analysis, some studies~\cite{yang2022contextual,yang2022disentangled} explored temporal correlations in videos and assigned the region of interest (RoI) accordingly. For example, motion vector and optical flow are adopted to implement tracking methods in~\cite{liu2019edge,guo2019distributed,mao2019catdet}. In~\cite{zhang2021elf}, the authors proposed an LSTM-based model to predict the RoI.
As for the multi-camera scenarios, many researchers exploited the physical correlations between cameras. For instance, in~\cite{jain2020spatula,jain2019scaling}, a location model was adopted to track target objects in a multi-camera network; Caesar~\cite{liu2019caesar} detected complex behaviors using the spatial-temporal correlations between surveillance cameras. Meanwhile, with the development of vehicle-to-everything (V2X) communication technology~\cite{yang2022novel}, the concept of cooperative perception~\cite{chen2019cooper,wang2020v2vnet,yang2022emotion,yang2022learning} was proposed to improve the vehicles' ability to perceive the system wide traffic conditions.
%augment the vehicles' system-wide perception of traffic conditions, 
Briefly, vehicles can share the self-acquired traffic information with other participants (e.g., other vehicles and traffic monitoring equipment). 
Different from the existing point cloud-based methods~\cite{hu2022where2comm,zhang2021emp,qiu2021autocast,xu2022v2xvit,li2021learning}, we propose utilizing multi-view video data from different vehicles to achieve more efficient cooperative perception. We need to solve the following challenges.

% \vspace{-0.5cm}
\begin{figure}[t]
    \centering
    \centerline{\includegraphics[width=0.4\textwidth]{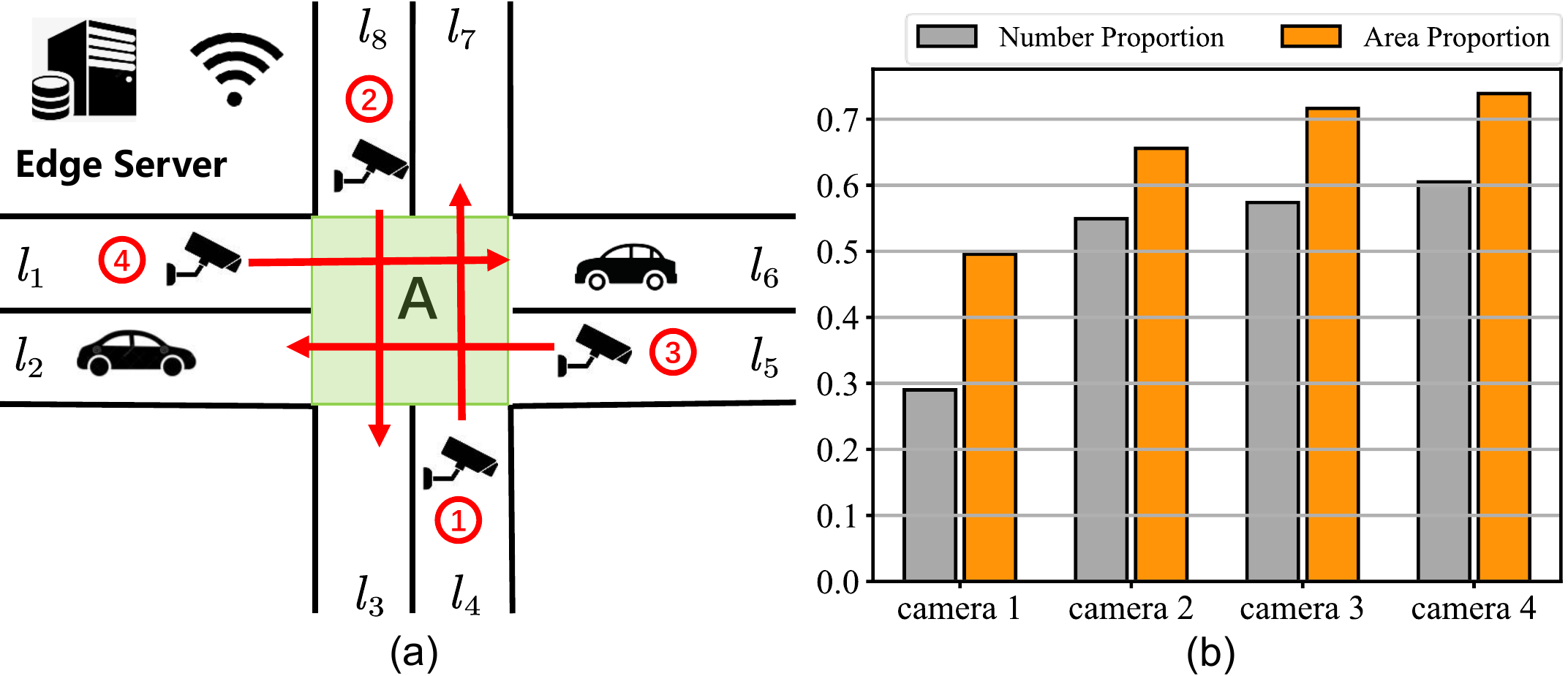}}
    \vspace{-0.5cm}
    \caption{Intersection and quantification of spatial redundancy.}
    \label{fig_intersection}
    \vspace{-15pt}
\end{figure}
% \vspace{-0.3cm}

% distribution of four cameras in \textit{AI City Challenge} dataset~\cite{Naphade21AIC21}

Fig.~\ref{fig_intersection}(a) shows a common method for deploying cameras at intersections. Since all four cameras can capture the vehicles in the intersection (region $A$), the video information about region $A$ is redundant for video analytics. Fig.~\ref{fig_intersection}(b) lists the number and area proportion of the vehicles located in the region $A$ among all detectable vehicles. The results indicate that lots of redundant information exist in the multi-view video data due to the overlapping detection ranges of cameras, namely spatial redundancy.
Moreover, considering that the correlations between consecutive frames produce another type of redundant information~\cite{kang2017noscope,li2020reducto,yang2023target} (called temporal redundancy), a mechanism that can eliminate these two types of redundancy simultaneously is needed. The last challenge is related to computing efficiency. Since the content of each video frame varies, we need to select the detection methods (e.g., YOLO~\cite{redmon2016you} and SSD~\cite{liu2016ssd}) adaptively rather than using a fixed detector.

To tackle the above challenges, we design a traffic-related object detection framework, CEVAS, to support video analytics-based cooperative perception. Summarily, we propose a fine-grained input filtering policy to produce the RoI from images, which can eliminate the temporal redundancy between consecutive frames and the spatial redundancy across cameras. Also, we adopt the sharing object manager to manage the objects with spatial redundancy and guarantee the detection accuracy of each agent through result sharing. Further, a lightweight model selection policy is introduced to select detection methods adaptively. 

% In the following sections, we will introduce each component of our model in detail.

% The main contributions are: 1) We propose a fine-grained input filtering policy to produce the RoI from images, eliminating the temporal redundancy between consecutive frames and the spatial redundancy across cameras. 2) We implement the sharing object manager to manage the objects with spatial redundancy and guarantee the detection accuracy of each agent through result sharing. 3) A lightweight model selection policy is introduced to select detection models adaptively. 4) Experimental results show that CEVAS significantly reduces the response latency while achieving the same detection accuracy compared to state-of-the-art (SOTA) methods.

\begin{figure}[t]
\centering
\subfigure[Traffic monitor.]{
\centering
\includegraphics[width=0.45\linewidth]{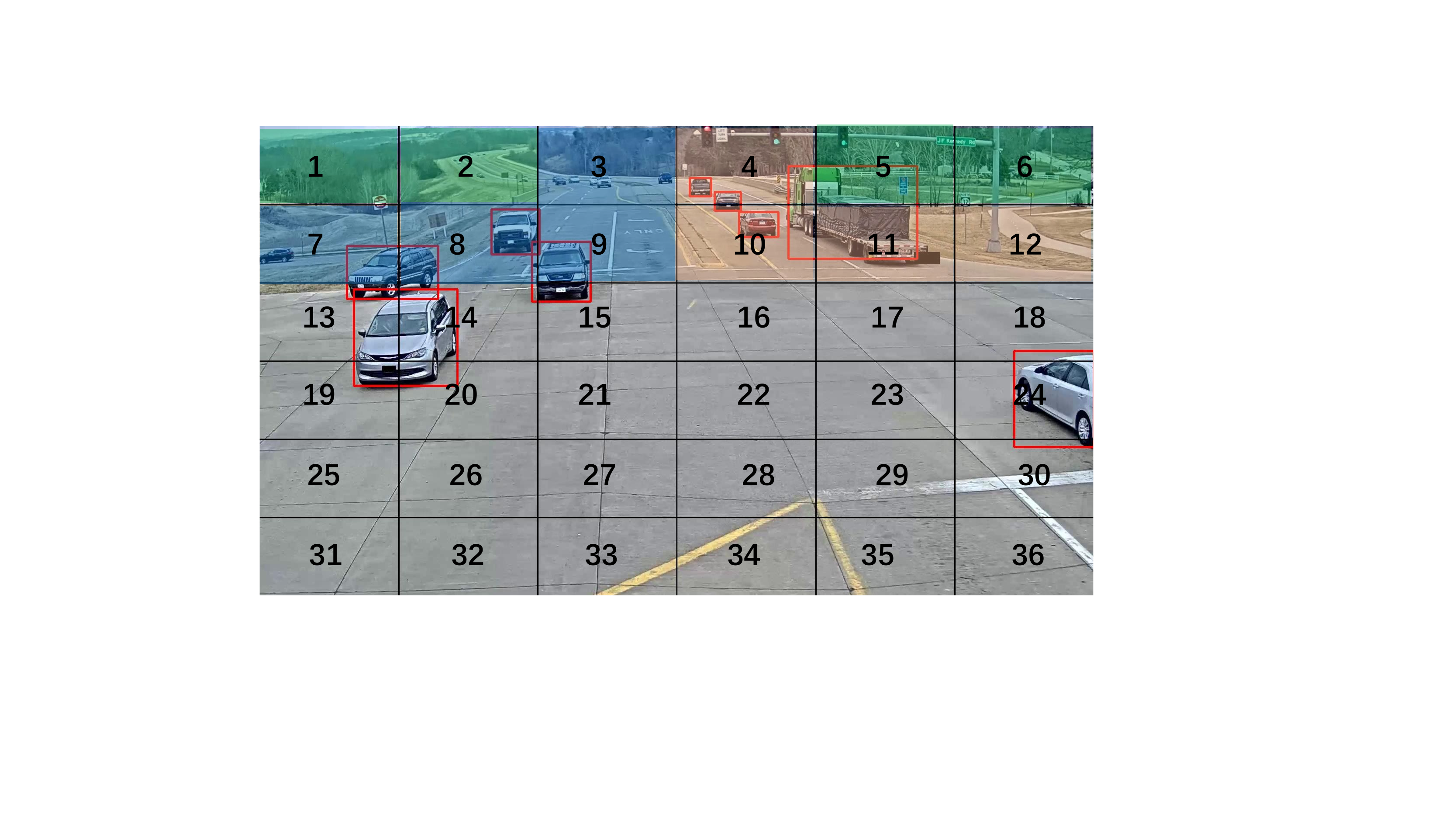}
\label{fig: camera3}
% \caption{fig_camera}
}
% \quad
% \hfill
% \subfigure[The captured image of traffic monitor $C_4$.]{
% \centering
% \includegraphics[width=0.28\linewidth]{imgs/camera4.pdf}
% \label{fig: camera4}
% }
\hfill
\subfigure[Vehicular camera.]{
\centering
\includegraphics[width=0.45\linewidth]{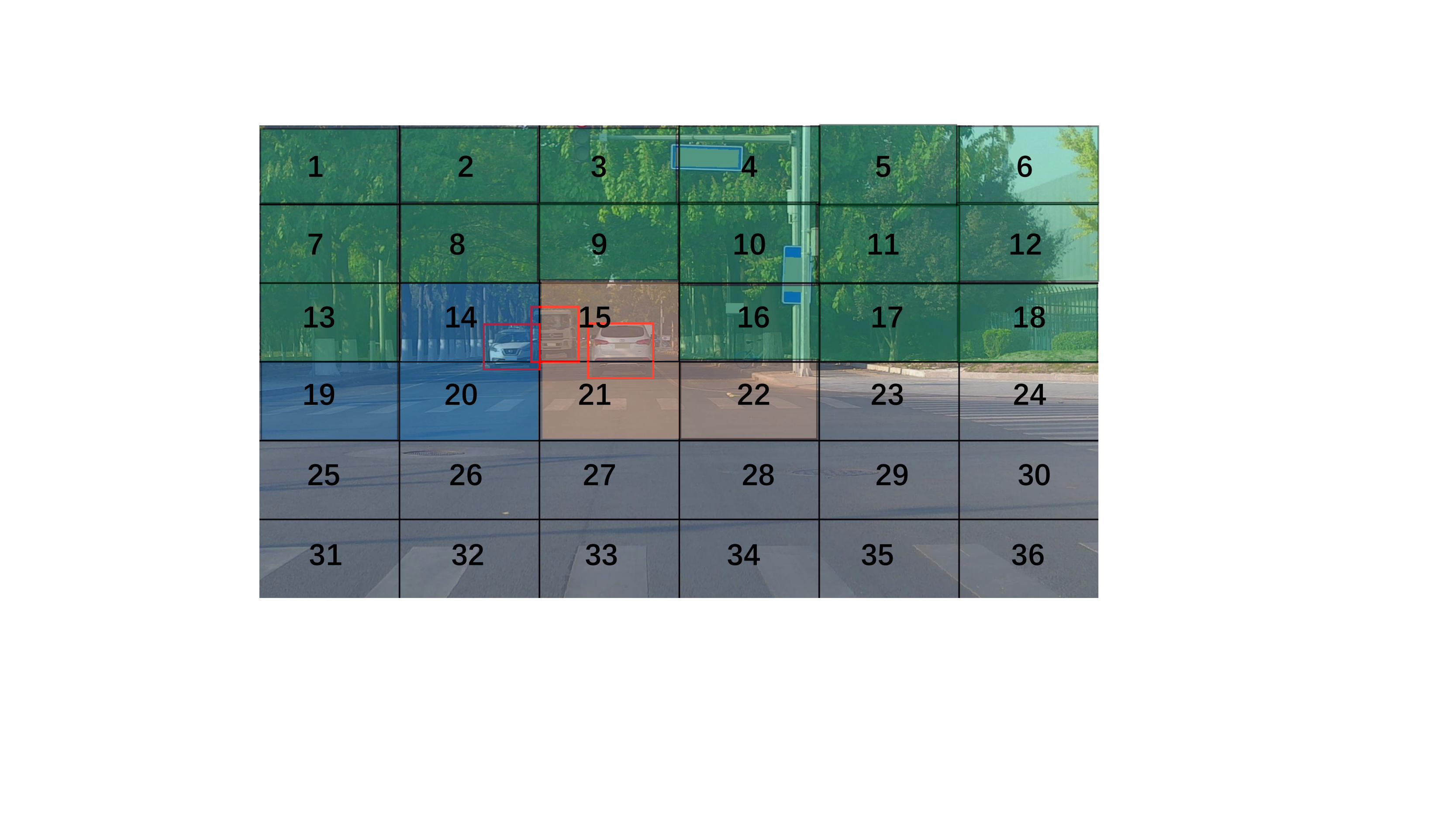}
\label{fig: camera5}
% \caption{fig_camera}
}
\vspace{-0.5cm}
\caption{The captured images of diverse agents.}
\vspace{-0.3cm}
\label{fig_view}
\end{figure}

\vspace{-0.3cm}
\section{System Model}
\vspace{-0.3cm}
\label{sec:model}
In this section, we will introduce our cooperative perception model and the corresponding region partitioning policy.
% divide the camera views into diverse regions based on the physical correlations between cameras.

\vspace{-0.3cm}
\subsection{Cooperative Perception Model}
\vspace{-0.2cm}
We consider the cooperative perception model in Fig.~\ref{fig_intersection}(a). $C_i$ denotes the $i$-th agent with cameras, including traffic monitors and vehicular cameras. The image captured by $C_i$ at time $t$ is denoted as $F_t^i$. Upon receiving $F_t^i$ via wireless connections, the edge server detect the objects $\{O^i_{t,j}, O^i_{t,j'}\}$ via detection methods and return results $D^i_t$, which contains the corresponding bounding boxes (bboxes) $\{p^i_{t,j}, p^i_{t,j'}\}$. 
% Due to the overlapping detection range, different agents may capture the same objects. For example, among the object set $\{O^i_{t,j}, O^i_{t,j'}\}$ and $\{O^k_{t,j}, O^k_{t,j'}\}$, the objects $O^i_{t,j}$ and $O^k_{t,j'}$ may correspond to the same object.
\vspace{-0.3cm}
\subsection{Region Partitioning Policy}
\vspace{-0.2cm}
% The images captured by traffic monitors and vehicular cameras in Fig. 2 are used as examples to introduce our region partitioning policy. 

We first divide the whole image into multiple blocks with a fixed size and number them sequentially. Let $b_k$ denote the blocks. %Then we set the following regions considering their positions.
Then, based on the location of the vehicles appearing in the picture and their driving direction, we divide the field of view of a camera into four regions as follows:

% The blocks where vehicles will not appear are set as the background region, marked in green in Fig.~\ref{fig_view}.
% The entrance lanes ($\{l_2,l_4,l_6,l_8\}$ in Fig.~\ref{fig_intersection}) correspond to the blue blocks in Fig.~\ref{fig_view}, which we regard as the incoming region, denoted as $R_{in}$. As the vehicles in $R_{in}$ gradually approach the intersection, they will occupy more pixels and be detected, treated as \textbf{new objects}. 
% The exit lanes ($\{l_1,l_3,l_5,l_7\}$ in Fig.~\ref{fig_intersection}) correspond with the orange blocks, treated as the leaving region $R_{l}$. The vehicles in $R_l$ will move away from the intersection and occupy fewer pixels.
% We regard the intersection (region $A$) as the overlapping region, denoted as $R_o$, corresponding to the blocks without colors in Fig.~\ref{fig_view}. The vehicles in $R_o$ are defined as \textbf{sharing objects}, which will be captured by multiple agents simultaneously despite the different driving states.
The blocks where vehicles will not appear are set as \emph{the background region}, marked in green in Fig.~\ref{fig_view}.
We define entrance lanes as \emph{the incoming region} $R_{in}$ and labeled as the blue blocks in Fig.~\ref{fig_view}. As the vehicles in $R_{in}$ gradually approach the intersection, they will occupy more pixels and be detected, treated as \textbf{new objects}. 
In contrast, the exit lanes is defined as \emph{the leaving region} $R_{l}$ and labeled as orange blocks in Fig.~\ref{fig_view}. The vehicles in $R_l$ will move away from the intersection and occupy fewer pixels.
We consider the intersection (region $A$) as the overlapping region $R_o$, corresponding to the blocks without colors in Fig.~\ref{fig_view}. The vehicles in $R_o$ are defined as \textbf{sharing objects}, which will be captured by multiple agents simultaneously despite the different driving states.

\begin{figure}[t]
    \centering
    \centerline{\includegraphics[width=0.42\textwidth]{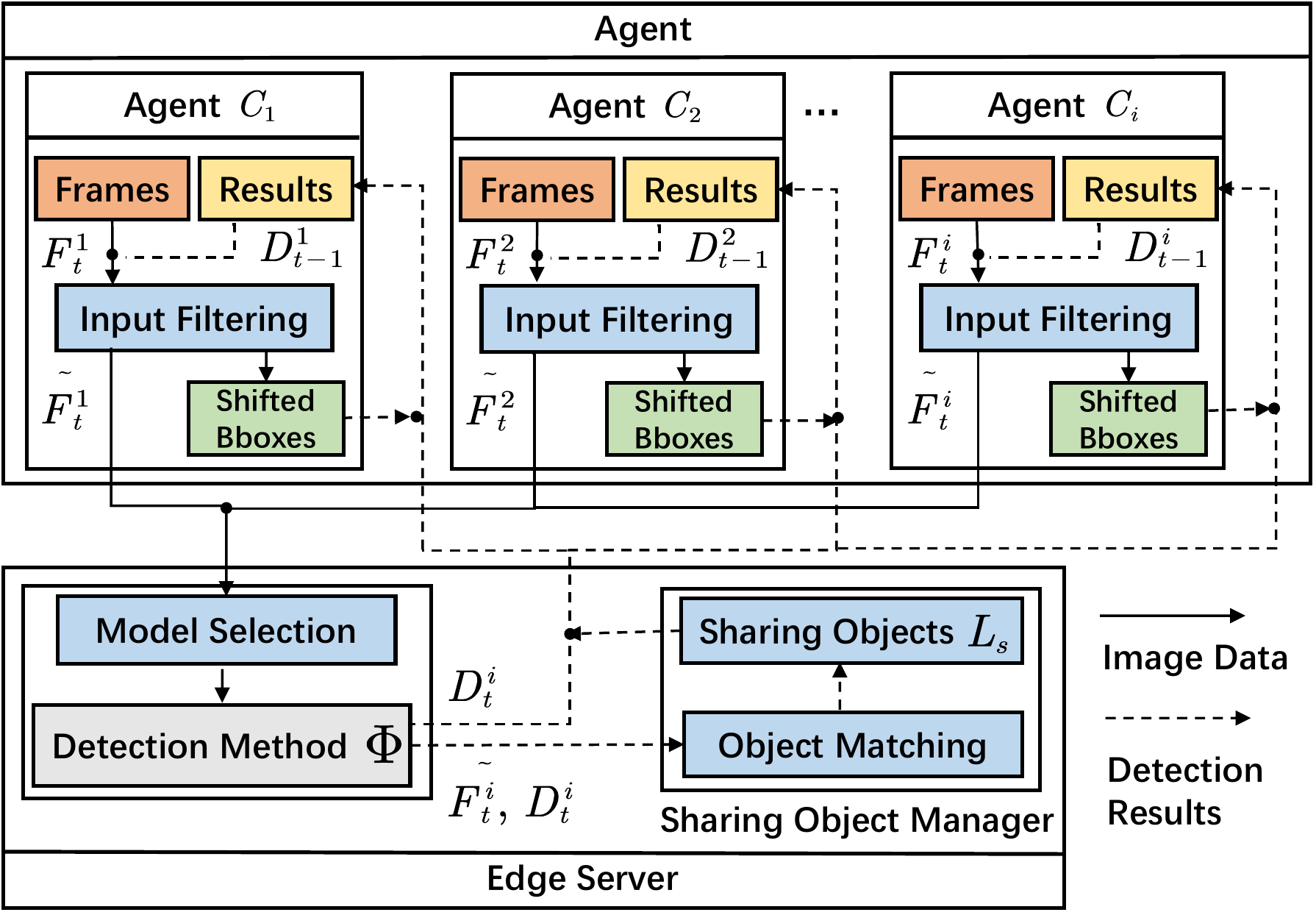}}
    \caption{The overview of our framework CEVAS.}
    \label{fig_overview}
    \vspace{-15pt}
\end{figure}

\vspace{-0.3cm}
\section{The proposed CEVAS framework} %Details of the proposed CEVAS framework
\vspace{-0.3cm}
%In this section, we will introduce our CEVAS framework in detail.
\label{sec:system}
\subsection{System Overview}
\vspace{-0.3cm}
Our framework CEVAS is divided into camera-side and server-side (see Fig.~\ref{fig_overview}). Agent $C_i$ first filters the current frame $F^i_t$ using the input filtering policy and then offloads the filtered frame $\widetilde{F^i_t}$ to the edge server. The edge server selects detection methods based on the frame content, then returns the detection results $D^i_t$ and the results of sharing objects to $C_i$. The sharing object manager works at the event-trigger pattern on the server side, analyzing the detection results to eliminate the spatial redundancy caused by sharing objects.

\vspace{-0.3cm}
\subsection{Input Filtering Policy}
\vspace{-0.2cm}
This section presents the fine-grained input filtering policy for producing a reasonable RoI (see Alg.~\ref{alg1}).
Different from the existing tracking-based methods~\cite{liu2019edge,guo2019distributed}, we use the motion information obtained from optical flow to predict the appearance of new objects. Moreover, benefiting from the region partitioning policy, we only need to process part of the area on images, which significantly reduces the computation burden on the camera side. We create the queue $Q_{off}$ and $Q_r$ to store offloaded blocks and shifted bboxes, respectively.

% \textbf{Step 1:} 
For each block $b_k$ in the incoming and leaving region, we first check if new objects appear in $b_k$. The number of pixels with non-zero optical flow value in $b_k$ is set as $n_k$. If $n_{k}$ is greater than the threshold $T_{new}$, we add $b_k$ into the queue $Q_{off}$. Then, as for each bbox $p_j$ in $D^i_{t-1}$, we calculate its motion offset $x_j$ and $y_j$ based on the optical flow, and use function $\Psi_{dis}(\cdot)$ to shift $p_j$ to $p'_{j}$ and obtain motion distance $d_j$, formulating as follows:
\begin{align}
    p'_j, d_j = \Psi_{dis}(p_j, x_j, y_j, f_d),
    \label{pre_1}\\
    d_j = f_d(x_j, y_j) = \sqrt{{x_j}^2+{y_j}^2},
\end{align}
where $f_{d}(\cdot)$ is the function to calculate motion distance. We set a threshold $T_{dis}$. If $d_j > T_{dis}$, we insert the corresponding blocks of $p'_{i}$ into $Q_{off}$. In contrast, we add $p'_{i}$ into $Q_r$ as the result of this object in the current frame since the object is probably at rest (lines 8-11 in Alg.~\ref{alg1}). Meanwhile, if $p'_{i}$ intersects with the overlapping region $R_o$, we also insert it into $Q_{off}$ since we need to share its information with other agents (line 8-9 in Alg.~\ref{alg1}).
In the end, we use $Q_{off}$ to obtain the filtered image $\widetilde{F^i_t}$ and upload $\widetilde{F^i_t}$ to the edge server.

% \textbf{Step 2:} $D^i_{t-1}$ denotes the detection results of previous frame $F^i_{t-1}$. 
% As for each bbox $p_j$ in $D^i_{t-1}$, we calculate its motion offset based on optical flow and shift it to get $p'_{i}$ (lines 7-8 in Alg.~\ref{alg1}). We set a motion threshold $T_{dis}$. If $d_j > T_{dis}$, we insert the corresponding blocks of $p'_{i}$ into $Q_{off}$. In contrast, we add the shifted bbox $p'_{i}$ into $Q_r$ as the result of this object in the current frame since the object is probably at rest (lines 9-12 in Alg.~\ref{alg1}). Meanwhile, suppose $p'_{i}$ intersects with the overlapping region $R_o$. In that case, we also insert it into $Q_{off}$ since we need to share its information with other agents (line 9-10 in Alg.~\ref{alg1}).
% In the end, we upload $Q_{off}$ to the edge server.      

\IncMargin{1em}
\begin{algorithm}[t]
\small
\SetKwData{Left}{left}\SetKwData{This}{this}\SetKwData{Up}{up} \SetKwFunction{Union}{Union}\SetKwFunction{FindCompress}{FindCompress}
\SetKwInOut{Input}{Input}\SetKwInOut{Output}{Output}
% 	\Input{The results $D^i_{t-1}$ of the previous frame} 
	\Input{Detection results $D^i_{t-1}$ of previous frame} 
	 Compute optical flow of current frame and create the queue $Q_{off}$ and $Q_r$\;
	 \For{each block $b_k \in \{R_{in} \cup R_{l}\}$}{ 
	 	Set the number of pixels with non-zero optical flow in $b_k$ as $n_k$\; 
	 	% $n_{new} \gets n_i - n_{pre}$\;
	 	\If{$n_{k} > T_{new}$}{Insert $b_k$ into $Q_{off}$\;}
 	 } 
 	 \For{each bbox $p_j \in D_{t-1}^i$}{ 
        % $p'_j \gets$ shift $p_j$ based on the mean offset $x_{j}$, $y_{j}$\;
        % $d_j \gets \sqrt{{x_j}^2+{y_j}^2}$\;
        $p'_j, d_j \gets$ Eq.~\ref{pre_1}\;
        \If{$d_j > T_{dis}$ or ($p'_j$ intersects with $R_o$)}{Insert the correponding blocks of $p'_j$ into $Q_{off}$\;}
        \Else{Insert $p'_j$ into $Q_r$\;}
        } 
 	 Obtain $\widetilde{F^i_t}$ from $Q_{off}$ and upload $\widetilde{F^i_t}$ to the edge server\;
 	    \caption{Input Filtering Policy}
 	    \label{alg1} 
 	 \end{algorithm}
   \vspace{-0.5cm}
 \DecMargin{1em}
 
% \vspace{-0.6cm}
\subsection{Model Selection Policy \label{modelselection}}
\vspace{-0.2cm}

% \begin{table}[htbp]
%      \caption{The candidate detection methods.}
%      \centering
%     %  \footnotesize
%      \scriptsize
%      \begin{tabular}{cccccc}
%       \toprule
%         Model  & Detection Accuracy & Parameter Quantity & Processing Speed  \\
%       \midrule
%      Model-1 & Lowest & Minimum & Fastest  \\
%      Model-2 & Medium & Medium & Medium \\
%      Model-3 & Highest & Maximum & Slowest \\
%       \bottomrule
%      \end{tabular}
%      \label{table_1}
%     \end{table}
In this section, we design a content-aware model selection policy to enable the adaptive selection of detection methods, which achieves a balance between accuracy and response latency. Let $\Phi$ denote the detection function, we can get
\begin{equation}
    D^i_t = \Phi(\widetilde{F^i_t}).
\end{equation}
In practice, the system can adapt to many detection methods with various characteristics and performances, for example, $\{\Phi_1, \Phi_2, \Phi_3\}$. 
Suppose the previous frame's results $D^i_{t-1}$ contains the bboxes $\{p_j \ | 1 \leq j \leq N\}$, inspired by~\cite{guo2021crossroi}, we propose to calculate the average intersection over union (IoU) value of all bboxes pair in $D^i_{t-1}$, as follows: 
% Can each bbox pair be represented as a variable
\begin{equation}
    m^i_{t-1} = \frac{N \cdot (N-1)}{2}\sum^{N-1}_{j=1} \sum^N_{j'=j+1} \Psi_{iou}(p_j,p_{j'}),
\end{equation}
in which $\Psi_{iou}(\cdot)$ is the function to calculate the IoU between two bboxes. When $m^i_{t-1}$ is 0, we select the fastest detection methods. Moreover, when $m^i_{t-1}$ exceeds the threshold $T_{iou}$, we select the model with the largest number of parameters to achieve a higher detection accuracy. In other cases, we select those methods with moderate speed and parameter quantities.

\IncMargin{1em}
\begin{algorithm}[t]
\small
\SetKwData{Left}{left}\SetKwData{This}{this}\SetKwData{Up}{up}
% \small
\SetKwFunction{Union}{Union}\SetKwFunction{FindCompress}{FindCompress} \SetKwInOut{Input}{Input}\SetKwInOut{Output}{Output}
	\Input{Detection results $D^i_{t}$ of current frame $\widetilde{F^i_t}$, sharing object list $L_s$} 
	 \For{each bbox $p_j \in D_{t}^i$}{ 
            \If{$p_i$ do not intersect with $R_o$}{continue\;}
	 	$O_j $, $\chi^c_j $, $\chi^h_j $ $\gets$ Eq.~\ref{preprocess}\; 
	 	% $R_i,O_M \gets$ Match($O_j$, $\chi^c_j$, $\chi^h_j$)\;
            \For{each object $O_m$ in $L_s$}{ 
            $s_{j,m} \gets$ Eq.~\ref{dis}\;
            Insert $s_{j,m}$ into the created queue $Q_s$\;
	 	% Compare the color features and feature pyramid of $O_i$ and that of $O_j$\; 
	 	% Insert the sum $s_{i,j}$ of the two feature vectors' Euclidean distance into created queue $Q_s$\;
 	 } 
            $s_{min} \gets$ the minimum value in $Q_s$\;
            $O_M \gets$ the corresponding sharing object of $s_{min}$\;
	 	\If{$s_min < T_s$}{\label{lt} 
                    Update result of $O_M$ to $p_i$\;
            % Create the list $L_c$ for $O_i$ and insert $C_i$ into $L_c$\;
	 		}
	 	\Else{\label{et} 
	 	    Insert $O_j$ into the list $L_s$\;
	 	    % Remove the previous camera $C_{pre}$ from $L_c$}
       }
       }
   %  \SetKwFunction{FMain}{Match}
   %  \SetKwProg{Fn}{Function}{:}{}
   %  \Fn{\FMain{$O_j$, $\chi^c_j$, $\chi^h_j$}}{
   %  %   \emph{Create the queue $Q_s$\;
   %    \For{each object $O_m$ in $L_s$}{ 
   %          $s_{j,m} \gets$ Eq.~\ref{dis}\;
   %          Insert $s_{j,m}$ into the created queue $Q_s$\;
	 	% % Compare the color features and feature pyramid of $O_i$ and that of $O_j$\; 
	 	% % Insert the sum $s_{i,j}$ of the two feature vectors' Euclidean distance into created queue $Q_s$\;
 	 % } 
   %    $s_{min} \gets$ the minimum value in $Q_s$\;
   %    $O_M \gets$ the corresponding sharing object of $s_{min}$\;
   %    \If{$s_{min} < T_s$}{\textbf{return} ($True,O_M$)}
	  % \Else{\textbf{return} ($False,O_M$)}
   %  }
 	    \caption{Sharing Object Manager}
            \vspace{-0.1cm}
 	    \label{alg3} 
 	 \end{algorithm}
   \vspace{-0.5cm}
 \DecMargin{1em}

% \vspace{-0.1cm}
\subsection{Sharing Object Manager}
\vspace{-0.1cm}
We implement the sharing object manager within the edge server to update the positions of the sharing objects in overlapping region.
Meanwhile, by sharing the positions of these sharing objects, our framework can ensure that each agent accurately detects the objects in their captured images. Since the previous works~\cite{jain2020spatula,guo2021crossroi} did not consider the detection accuracy of each agent while eliminating redundancy, they can not be applied in cooperative perception.
The process in Alg.~\ref{alg3} will be triggered when obtaining the detection results $D^i_{t}$.

For each bbox $p_j$ in $D^i_t$, we first check if it intersects with the overlapping region, then use the function $\Psi_{feat}$ to obtain the corresponding object $O_j$ and its features, as follows:
\begin{equation}
    O_j, \chi^c_j, \chi^h_j = \Psi_{feat}(p_j, \widetilde{F^i_t}, f_c, f_h),
    \label{preprocess}
\end{equation}
where $\chi^c_j \in \mathbb{R}^{1 \times 3}$ and $\chi^h_j \in \mathbb{R}^{C \times H \times W}$ denotes the color features and image features of the object $O_j$, respectively. $f_c$ and $f_e$ denotes two feature extractors, formulating as follows:
\begin{align}
    \chi^c_j = f_c(p_j, \widetilde{F^i_t}; \theta_c), \ \chi^c_j \in \mathbb{R}^{1 \times 3},\\
    \chi^h_j = f_h(p_j, \widetilde{F^i_t}; \theta_h), \chi^h_j \in \mathbb{R}^{C \times H \times W}.
\end{align}
Then, we match the object $O_j$ with the existing sharing objects in the list $L_c$. We obtain the distance $s_{j,m}$ between the features of $O_j$ and $O_m$ by
\begin{equation}
    s_{j,m} = \sqrt{\sum^{N_1}_{i=1} {(\chi^c_{j}(i)-\chi^c_{m}(i))}^2} + \sqrt{\sum^{N_2}_{i=1} {(\chi^h_{j}(i)-\chi^h_{m}(i))}^2}.
    \label{dis}
\end{equation}
We then insert $s_{j,m}$ into the queue $Q_s$. After iteration, the minimum value in $Q_s$ is set as $s_{min}$, and the corresponding sharing object is $O_M$. If $s_{min}$ is less than the threshold $T_s$, $O_j$ is considered to be the same object with $O_M$, so we update the detection results of $O_M$ to $p_i$.
In contrast, we add $O_j$ to $L_s$ as a new sharing object.
The results of the sharing objects in $L_s$ will be shared among all agents.

% \textbf{Step 1:}  
% After cropping the object $O_i$ from the original image based on the bbox $p_i$, we use bilinear interpolation to resize $O_i$ to a fixed size and extract its color features. Also, we use autoencoder to extract the feature pyramid of $O_i$ for a higher matching accuracy.

% \textbf{Step 2:} 
% Here, we call the $Match$ function to match the object $O_i$ with the existing sharing objects. 
% We calculate the Euclidean distances between the color features and feature pyramid of $O_i$ and that of existing sharing objects, then insert the sum $s_{i,j}$ of Euclidean distances into the queue $Q_s$. The minimum value in $Q_s$ is set as $s_{min}$, and the corresponding sharing object is $O_M$. If $s_{min}$ is less than the threshold $T_s$, $O_i$ is considered to be the same object with $O_M$.

% \textbf{Step 3:} 
% The list $L_s$ contains the existing sharing objects. 
% If the matching result $R_i$ is $False$, we add $O_i$ to $L_s$ as a new sharing object and record the corresponding agent $C_i$ into the list $L_c$ (lines 5-6). If successfully matched, we will update the detection results of $O_M$ to $p_i$ and insert $C_{i}$ to $L_c$.
% The results of the sharing objects in $L_s$ will be shared among all agents.

% \begin{figure*}[t]
%     \centering
%     \centerline{\includesvg[width=0.95\textwidth]{single_results/frame-period-new.svg}}
%     \vspace{-0.5cm}
%     \caption{Impact of Frame Interval.}
%     \label{fig_frame}
%     \vspace{-0.3cm}
% \end{figure*}

\begin{figure*}[t]
    \centering
    \centerline{\includegraphics[width=0.95\textwidth]{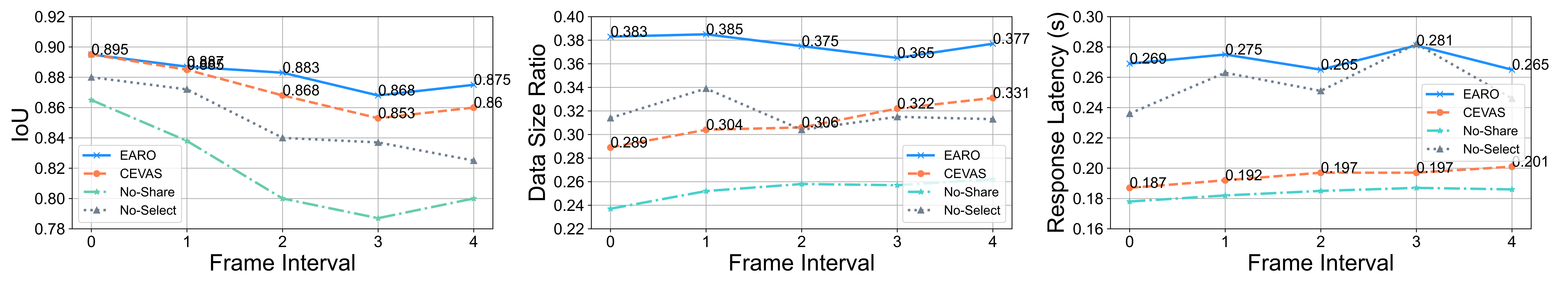}}
    \vspace{-0.5cm}
    \caption{Impact of Frame Interval.}
    \label{fig_frame}
    \vspace{-0.3cm}
\end{figure*}

\begin{figure*}[t]
    \centering
    \centerline{\includegraphics[width=0.95\textwidth]{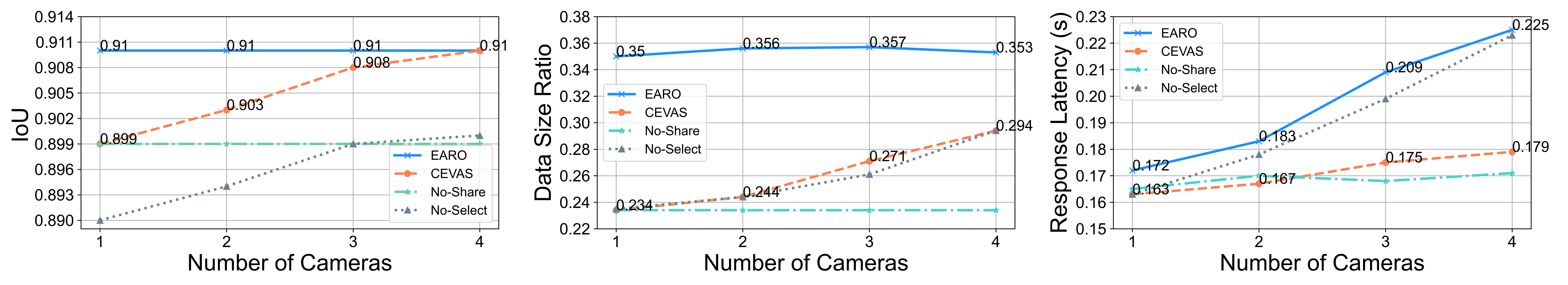}}
    \vspace{-0.5cm}
    \caption{Impact of the Number of Cameras.}
    %  For better visualization and comparison, only the values of our proposed CEVAS and SOTA method.
    \label{fig_camera}
    \vspace{-0.5cm}
\end{figure*}

% \begin{figure*}[t]
%     \centering
%     \centerline{\includesvg[width=0.95\textwidth]{single_results/camera-new.svg}}
%     \vspace{-0.5cm}
%     \caption{Impact of the Number of Cameras.}
%     %  For better visualization and comparison, only the values of our proposed CEVAS and SOTA method.
%     \label{fig_camera}
%     \vspace{-0.5cm}
% \end{figure*}

% \begin{figure*}[t]
% \centering
% \subfigure{
% \centering
% \includesvg[width=0.3\textwidth]{single_results/frame-period(2)-1.svg}
% \label{fig_f1_3}
% }
% % \quad
% \hfill
% \subfigure{
% \centering
% \includesvg[width=0.3\textwidth]{single_results/frame-period(2).svg}
% \label{fig_f1_2}
% }
% \hfill
% \subfigure{
% \centering
% \includesvg[width=0.3\textwidth]{single_results/frame-period(1)-1.svg}
% \label{fig_f1_1}
% }
% \vspace{-0.3cm}
% \caption{Impact of Frame Interval.}
% \label{fig_frame}
% \vspace{-0.3cm}
% \end{figure*}

% \vspace{-0.3cm}
% \begin{figure*}[!htbp]
% \centering
% \subfigure{
% \centering
% \includesvg[width=0.3\textwidth]{single_results/camera(2)-1.svg}
% \label{fig_f2_1}}
% % \quad
% \hfill
% \subfigure{
% \centering
% \includesvg[width=0.3\textwidth]{single_results/camera(1)-1.svg}
% \label{fig_f2_2}
% }
% \hfill
% \subfigure{
% \centering
% \includesvg[width=0.3\textwidth]{single_results/camera(2).svg}
% \label{fig_f2_3}
% }
% \vspace{-0.3cm}
% \caption{Impact of the number of cameras.}
% \label{fig_camera}
% \vspace{-0.3cm}
% \end{figure*}

\vspace{-0.3cm}
\section{PERFORMANCE EVALUATION}
\vspace{-0.3cm}
% In this section, we conduct extensive simulations on our platform\footnote{https://github.com/bruceteams/CEVAS} to evaluate the performance of CEVAS. 
In this section, we conduct extensive simulations on our platform~\cite{CEVAS} to evaluate the performance of CEVAS.
\vspace{-0.3cm}
\subsection{Implementation Details}
\vspace{-0.2cm}
\textbf{Simulation Settings.} 
From \textit{AI city challenge} dataset~\cite{Naphade21AIC21}, we select a video clip of an intersection in which every four camera captures video at 10 Hz. Three variants of YOLOv5 models~\cite{yolov5} with different speeds and accuracy are set as the corresponding detection methods. We set $T_{dis}$ as 0.1, $T_{iou}$ as 0.2, $T_s$ as 0.05. $T_{new}$ is set as a quarter of the number of pixels in a block. 

\textbf{Compared Schemes.}
1) EARO~\cite{liu2019edge}: EARO implements a motion vector-based object tracking mechanism to assign RoI and compress the image data of other areas. 
2) No-Share: We disable the sharing object manager in this scheme, so the edge server will not share the detection results of sharing objects with other agents.
3) No-Select: This scheme is an ablation study with the model selection policy. The edge server will process the received images using a fixed detection method.

\textbf{Evaluation Metrics.}
1) IoU: IoU is used to evaluate the detection accuracy between the predicted bboxes and ground truth.
2) Data Size Ratio: The ratio of the amount of offloaded data to the amount of original image data.
3) Response Latency: The time consumed to detect a frame, including on-camera filtering, network transmission, and inference latency.
\vspace{-0.8cm}
\subsection{Impact of Frame Interval}
\vspace{-0.2cm}
We simulate the camera with different \textit{fps} by setting the interval between two consecutive frames. When the frame interval is $n$, we pick one frame from every $n$ frames. 
As shown in Fig.~\ref{fig_frame}, it is apparent that CEVAS reduces the data size ratio and response latency compared to EARO, with a decrease of 18\% and 28\%, respectively. This indicates that our proposed input filtering policy significantly eliminates spatial and temporal redundancy, avoiding the transmission of redundant information. Moreover, CEVAS greatly reduces response latency compared with No-Select. The potential reason is that our model selection policy enables the adaptive selection of detection methods and thus increases computing efficiency.   
Meanwhile, CEVAS only decreases 1\% in IoU compared to EARO and improves IoU by 7\% and 2.5\% compared to No-Share and No-Select. 
The above results show that CEVAS maintains high detection accuracy for each agent using fewer image data. 

% \vspace{-0.3cm}
% \begin{figure}[t]
% \centering
% \subfigure[Impact of Frame Interval on Data Size Ratio.]{
% \centering
% \includesvg[width=0.23\textwidth]{single_results/frame-period(1)-1.svg}
% \label{fig_f1_1}
% }
% % \quad
% \subfigure[Impact of Frame Interval on Response Latency.]{
% \centering
% \includesvg[width=0.21\textwidth]{single_results/frame-period(2).svg}
% \label{fig_f1_2}
% }
% \quad
% \subfigure[Impact of Frame Interval on IoU.]{
% \centering
% \includesvg[width=0.22\textwidth]{single_results/frame-period(2)-1.svg}
% \label{fig_f1_3}
% }
% % \quad
% \subfigure[Impact of the number of cameras on IoU.]{
% \centering
% \includesvg[width=0.22\textwidth]{single_results/camera(2)-1.svg}
% \label{fig_f2_1}}
% \quad
% \subfigure[Impact of the number of cameras on Data Size Ratio.]{
% \centering
% \includesvg[width=0.23\textwidth]{single_results/camera(1)-1.svg}
% \label{fig_f2_2}
% }
% \subfigure[Impact of the number of cameras on Response Latency.]{
% \centering
% \includesvg[width=0.21\textwidth]{single_results/camera(2).svg}
% \label{fig_f2_3}
% }
% \vspace{-0.3cm}
% \caption{Quantitative Experiments.}
% \vspace{-0.3cm}
% \end{figure}

\vspace{-0.2cm}
\subsection{Impact of the Number of Cameras}
\vspace{-0.2cm}
To intuitively indicate our exploitation of spatial redundancy, we investigate the impact of the number of cameras. 
The corresponding results are shown in Fig.~\ref{fig_camera}. As the number of cameras increases, the methods that utilize spatial redundancy using the sharing object manager, including CEVAS and No-Select, achieve a more accurate detection, and the IoU of CEVAS gradually approaches EARO. Meanwhile, CEVAS achieves a 26\% reduction in average data size ratio and 13\% reduction in average response latency compared to EARO. 
Additionally, the response latency of EARO and No-Select without the model selection policy rises linearly with the number of cameras. In contrast, CEVAS maintains lower response latency as it avoids the unnecessary processing of redundant information and uses detection methods adaptively. From the above results, we can see that our framework can support multiple cameras and ensure scalability in large-scale environments.

\vspace{-0.2cm}
\subsection{Impact of Transmission Rate}
\vspace{-0.2cm}
Lastly, we explore the impact of the transmission rate of agents on response latency. As Table~\ref{tab:t1-bandwidth} shows, CEVAS reduces the response latency by 31\% on average compared with EARO. It indicates that CEVAS accelerates video analytics by eliminating spatial-temporal redundancy and selecting detection methods adaptively.

\vspace{-0.3cm}

\begin{table}[t]
    \centering
    \caption{Impact of transmission rate on response latency.}
    \setlength{\tabcolsep}{3.2mm}{
    % \small
    \label{tab:t1-bandwidth}
    \resizebox{0.47\textwidth}{!}{
        \begin{tabular}{|c|ccccc|}%@{}cccccc@{}
\toprule  

% \multirow{2}{*}{\textbf{\diagbox{Method}{latency}{\# Rate}}} & \multicolumn{5}{c}{\textbf{Response Latency (s)}} \\ \cmidrule(l){2-6} 
           \textbf{\diagbox{Method}{Latency (s)}{\# Rate}}                           & 80       & 100      & 120     & 140     & 160     \\ \midrule

EARO \cite{liu2019edge}                                 & 0.269    & 0.263    & 0.254   & 0.242   & 0.227   \\
CEVAS (Ours)                                & \textbf{0.187} & \textbf{0.183} & \textbf{0.175} & \textbf{0.166} & \textbf{0.155} \\ \bottomrule
\end{tabular}}}
\vspace{-0.4cm}
\end{table}

\vspace{-0.2cm}
\section{CONCLUSION}
\vspace{-0.2cm}
In this paper, to achieve efficient cooperative perception, we proposed a traffic-related object detection framework CEVAS, which simultaneously eliminates the existing spatial and temporal redundancy in multi-view video data. 
Extensive experimental results demonstrated that our framework could considerably reduce the response latency while ensuring the detection accuracy of each agent.
% We also implemented a content-aware model selection policy to select the detection methods adaptively. Extensive experimental results demonstrated that our framework could considerably reduce the response latency while ensuring the detection accuracy of each agent.

\vfill\pagebreak

% \section{REFERENCES}
% \label{sec:refs}
% References should be produced using the bibtex program from suitable
% BiBTeX files (here: strings, refs, manuals). The IEEEbib.bst bibliography
% style file from IEEE produces unsorted bibliography list.
% -------------------------------------------------------------------------
\footnotesize
\bibliographystyle{IEEEbib}
\bibliography{main}

\begin{thebibliography}{10}

\bibitem{guo2021crossroi}
Hongpeng Guo, Shuochao Yao, Zhe Yang, Qian Zhou, and Klara Nahrstedt,
\newblock ``Crossroi: Cross-camera region of interest optimization for
  efficient real time video analytics at scale,''
\newblock in {\em Proc. ACM MMSys}, 2021, pp. 186--199.

\bibitem{liu2022appearance}
Yang Liu, Jing Liu, Jieyu Lin, Mengyang Zhao, and Liang Song,
\newblock ``Appearance-motion united auto-encoder framework for video anomaly
  detection,''
\newblock {\em IEEE Transactions on Circuits and Systems II: Express Briefs},
  vol. 69, no. 5, pp. 2498--2502, 2022.

\bibitem{liu2022collaborative}
Yang Liu, Jing Liu, Mengyang Zhao, Shuang Li, and Liang Song,
\newblock ``Collaborative normality learning framework for weakly supervised
  video anomaly detection,''
\newblock {\em IEEE Transactions on Circuits and Systems II: Express Briefs},
  vol. 69, no. 5, pp. 2508--2512, 2022.

\bibitem{jain2020spatula}
Samvit Jain, Xun Zhang, Yuhao Zhou, Ganesh Ananthanarayanan, Junchen Jiang,
  Yuanchao Shu, Paramvir Bahl, and Joseph Gonzalez,
\newblock ``Spatula: Efficient cross-camera video analytics on large camera
  networks,''
\newblock in {\em Proc. IEEE/ACM SEC}, 2020, pp. 110--124.

\bibitem{liu2022learning}
Yang Liu, Jing Liu, Mengyang Zhao, Dingkang Yang, Xiaoguang Zhu, and Liang
  Song,
\newblock ``Learning appearance-motion normality for video anomaly detection,''
\newblock in {\em 2022 IEEE International Conference on Multimedia and Expo
  (ICME)}. IEEE, 2022, pp. 1--6.

\bibitem{yang2022contextual}
Dingkang Yang, Shuai Huang, Yang Liu, and Lihua Zhang,
\newblock ``Contextual and cross-modal interaction for multi-modal speech
  emotion recognition,''
\newblock {\em IEEE Signal Processing Letters}, vol. 29, pp. 2093--2097, 2022.

\bibitem{yang2022disentangled}
Dingkang Yang, Shuai Huang, Haopeng Kuang, Yangtao Du, and Lihua Zhang,
\newblock ``Disentangled representation learning for multimodal emotion
  recognition,''
\newblock in {\em Proceedings of the 30th ACM International Conference on
  Multimedia}, 2022, p. 1642–1651.

\bibitem{liu2019edge}
Luyang Liu, Hongyu Li, and Marco Gruteser,
\newblock ``Edge {{Assisted Real-time Object Detection}} for {{Mobile Augmented
  Reality}},''
\newblock in {\em Proc. ACM MobiCom}, 2019, pp. 1--16.

\bibitem{guo2019distributed}
Yundi Guo, Beiji Zou, Ju~Ren, Qingqing Liu, Deyu Zhang, and Yaoxue Zhang,
\newblock ``Distributed and efficient object detection via interactions among
  devices, edge, and cloud,''
\newblock {\em IEEE Trans. Multimedia}, vol. 21, no. 11, pp. 2903--2915, 2019.

\bibitem{mao2019catdet}
Huizi Mao, Taeyoung Kong, et~al.,
\newblock ``Catdet: Cascaded tracked detector for efficient object detection
  from video,''
\newblock {\em Proc. MLSys}, vol. 1, pp. 201--211, 2019.

\bibitem{zhang2021elf}
Wuyang Zhang, Zhezhi He, Luyang Liu, Zhenhua Jia, Yunxin Liu, Marco Gruteser,
  Dipankar Raychaudhuri, and Yanyong Zhang,
\newblock ``Elf: Accelerate high-resolution mobile deep vision with
  content-aware parallel offloading,''
\newblock in {\em Proc. ACM MobiCom}, 2021, pp. 201--214.

\bibitem{jain2019scaling}
Samvit Jain, Ganesh Ananthanarayanan, Junchen Jiang, Yuanchao Shu, and Joseph
  Gonzalez,
\newblock ``Scaling {{Video Analytics Systems}} to {{Large Camera
  Deployments}},''
\newblock in {\em Proc. HotMobile}, 2019, pp. 9--14.

\bibitem{liu2019caesar}
Xiaochen Liu, Pradipta Ghosh, Oytun Ulutan, B.~S. Manjunath, Kevin Chan, and
  Ramesh Govindan,
\newblock ``Caesar: Cross-camera complex activity recognition,''
\newblock in {\em Proc. ACM SenSys}, 2019, pp. 232--244.

\bibitem{yang2022novel}
Kun Yang, Peng Sun, Jieyu Lin, Azzedine Boukerche, and Liang Song,
\newblock ``A novel distributed task scheduling framework for supporting
  vehicular edge intelligence,''
\newblock in {\em 2022 IEEE 42nd International Conference on Distributed
  Computing Systems (ICDCS)}. IEEE, 2022, pp. 972--982.

\bibitem{chen2019cooper}
Qi~Chen, Sihai Tang, Qing Yang, and Song Fu,
\newblock ``Cooper: Cooperative perception for connected autonomous vehicles
  based on 3d point clouds,''
\newblock in {\em Proc. ICDCS}, 2019, pp. 514--524.

\bibitem{wang2020v2vnet}
Tsun-Hsuan Wang, Sivabalan Manivasagam, Ming Liang, Bin Yang, Wenyuan Zeng, and
  Raquel Urtasun,
\newblock ``V2vnet: Vehicle-to-vehicle communication for joint perception and
  prediction,''
\newblock in {\em Proc. ECCV}, 2020, pp. 605--621.

\bibitem{yang2022emotion}
Dingkang Yang, Shuai Huang, Shunli Wang, Yang Liu, Peng Zhai, Liuzhen Su,
  Mingcheng Li, and Lihua Zhang,
\newblock ``Emotion recognition for multiple context awareness,''
\newblock in {\em European Conference on Computer Vision}. 2022, vol. 13697,
  pp. 144--162, Springer.

\bibitem{yang2022learning}
Dingkang Yang, Haopeng Kuang, Shuai Huang, and Lihua Zhang,
\newblock ``Learning modality-specific and -agnostic representations for
  asynchronous multimodal language sequences,''
\newblock in {\em Proceedings of the 30th ACM International Conference on
  Multimedia}, 2022, p. 1708–1717.

\bibitem{hu2022where2comm}
Yue Hu, Shaoheng Fang, Zixing Lei, Yiqi Zhong, and Siheng Chen,
\newblock ``Where2comm: Communication-efficient collaborative perception via
  spatial confidence maps,''
\newblock {\em arXiv preprint arXiv:2209.12836}, 2022.

\bibitem{zhang2021emp}
Xumiao Zhang, Anlan Zhang, Jiachen Sun, Xiao Zhu, Y.~Ethan Guo, Feng Qian, and
  Z.~Morley Mao,
\newblock ``Emp: Edge-assisted multi-vehicle perception,''
\newblock in {\em Proc. ACM MobiCom}, 2021, pp. 545--558.

\bibitem{qiu2021autocast}
Hang Qiu, Po-Han Huang, Namo Asavisanu, Xiaochen Liu, Konstantinos Psounis, and
  Ramesh Govindan,
\newblock ``Autocast: Scalable infrastructure-less cooperative perception for
  distributed collaborative driving,''
\newblock in {\em Proc. ACM MobiSys}, 2022, p. 128–141.

\bibitem{xu2022v2xvit}
Runsheng Xu, Hao Xiang, Zhengzhong Tu, Xin Xia, Ming-Hsuan Yang, and Jiaqi Ma,
\newblock ``V2x-vit: Vehicle-to-everything cooperative perception with vision
  transformer,''
\newblock {\em arXiv preprint arXiv:2203.10638}, 2022.

\bibitem{li2021learning}
Yiming Li, Shunli Ren, Pengxiang Wu, Siheng Chen, Chen Feng, and Wenjun Zhang,
\newblock ``Learning distilled collaboration graph for multi-agent
  perception,''
\newblock in {\em Proc. NeurIPS}, 2021, vol.~34, pp. 29541--29552.

\bibitem{kang2017noscope}
Daniel Kang, John Emmons, Firas Abuzaid, Peter Bailis, and Matei Zaharia,
\newblock ``Noscope: Optimizing neural network queries over video at scale,''
\newblock {\em Proc. VLDB Endow.}, vol. 10, no. 11, pp. 1586–1597, 2017.

\bibitem{li2020reducto}
Yuanqi Li, Arthi Padmanabhan, Pengzhan Zhao, Yufei Wang, Guoqing~Harry Xu, and
  Ravi Netravali,
\newblock ``Reducto: On-camera filtering for resource-efficient real-time video
  analytics,''
\newblock in {\em Proc. ACM SIGCOMM}, 2020, pp. 359--376.

\bibitem{yang2023target}
Dingkang Yang, Yang Liu, Can Huang, Mingcheng Li, Xiao Zhao, Yuzheng Wang, Kun
  Yang, Yan Wang, Peng Zhai, and Lihua Zhang,
\newblock ``Target and source modality co-reinforcement for emotion
  understanding from asynchronous multimodal sequences,''
\newblock {\em Knowledge-Based Systems}, p. 110370, 2023.

\bibitem{redmon2016you}
Joseph Redmon, Santosh Divvala, Ross Girshick, and Ali Farhadi,
\newblock ``You only look once: Unified, real-time object detection,''
\newblock in {\em Proc. IEEE CVPR}, 2016, pp. 779--788.

\bibitem{liu2016ssd}
Wei Liu, Dragomir Anguelov, Dumitru Erhan, Christian Szegedy, Scott Reed,
  Cheng-Yang Fu, and Alexander~C Berg,
\newblock ``Ssd: Single shot multibox detector,''
\newblock in {\em Proc. ECCV}, 2016, pp. 21--37.

\bibitem{CEVAS}
``Cevas,'' https://github.com/bruceteams/CEVAS.

\bibitem{Naphade21AIC21}
Milind Naphade, Shuo Wang, David~C. Anastasiu, Zheng Tang, Ming-Ching Chang,
  Xiaodong Yang, Yue Yao, Liang Zheng, Pranamesh Chakraborty, Christian~E.
  Lopez, Anuj Sharma, Qi~Feng, Vitaly Ablavsky, and Stan Sclaroff,
\newblock ``The 5th {AI City Challenge},''
\newblock in {\em Proc. IEEE CVPRW}, 2021.

\bibitem{yolov5}
``Yolov5,'' https://pytorch.org/hub/ultralytics\_yolov5.

\end{thebibliography}
% \vspace{0.5cm}
\end{document}